\documentclass{article}

\usepackage[final,nonatbib]{neurips_2020}
\usepackage[numbers]{natbib}

\usepackage[utf8]{inputenc} 
\usepackage[T1]{fontenc}    
\usepackage{hyperref}       
\usepackage{url}            
\usepackage{booktabs}       
\usepackage{amsfonts}       
\usepackage{nicefrac}       
\usepackage{microtype}      

\usepackage{amsmath,amsfonts,bm}

\def\eqref#1{equation~\ref{#1}}

\def\1{\bm{1}}

\DeclareMathAlphabet{\mathsfit}{\encodingdefault}{\sfdefault}{m}{sl}
\SetMathAlphabet{\mathsfit}{bold}{\encodingdefault}{\sfdefault}{bx}{n}

\usepackage{amsmath}
\usepackage{amssymb}
\usepackage{natbib}
\usepackage{wrapfig}
\usepackage{multirow}
\usepackage[ruled,vlined]{algorithm2e}

\usepackage[font=small]{caption}
\usepackage{graphicx}
\graphicspath{{figures/}}

\usepackage{titlesec}
\titlespacing{\paragraph}{%
  0pt}{
  0em}{
  1.1em}

\title{Offline Learning from Demonstrations \\ and Unlabeled Experience}

\author{
Konrad \.Zo\l{}na$^{1}$
\And
Alexander Novikov$^{1}$
\And
Ksenia Konyushkova$^{1}$
\And
Caglar Gulcehre$^{1}$
\And
Ziyu Wang$^{2}$
\And
Yusuf Aytar$^{1}$
\And
Misha Denil$^{1}$
\And
Nando de Freitas$^{1}$
\And
Scott Reed$^{1}$
\AND
$^{1}${\normalfont DeepMind} \mbox{ } \mbox{ } \mbox{ } \mbox{ } \mbox{ }
$^{2}${\normalfont Google Brain}\\\\
\texttt{kondiz@google.com}
}

\begin{document}
\maketitle


\begin{abstract}
Behavior cloning (BC) is often practical for robot learning because it allows a policy to be trained offline without rewards, by supervised learning on expert demonstrations.
However, BC does not effectively leverage what we will refer to as \emph{unlabeled experience}: data of mixed and unknown quality without reward annotations.
This unlabeled data can be generated by a variety of sources such as human teleoperation, scripted policies and other agents on the same robot.
Towards data-driven offline robot learning that can use this unlabeled experience, we introduce Offline Reinforced Imitation Learning (ORIL).
ORIL first learns a reward function by contrasting observations from demonstrator and unlabeled trajectories, then annotates all data with the learned reward, and finally trains an agent via offline reinforcement learning.
Across a diverse set of continuous control and simulated robotic manipulation tasks, we show that ORIL consistently outperforms comparable BC agents by effectively leveraging unlabeled experience.
\end{abstract}

\section{Introduction}
\begin{wrapfigure}{r}{0.45\textwidth}
\vspace{-0.45cm}
  \centering
  \includegraphics[width=1\linewidth]{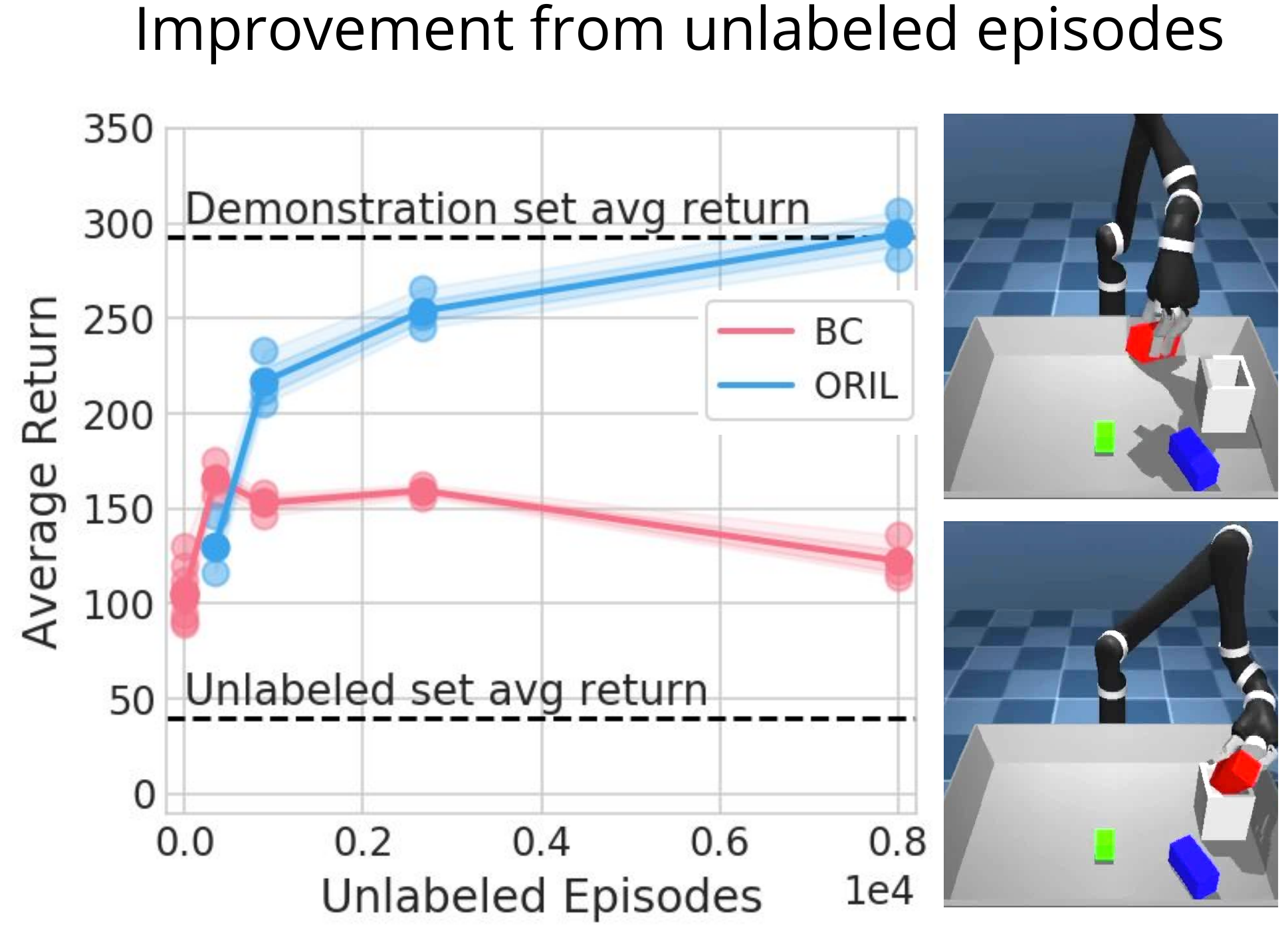}
  \caption{\textbf{Performance of ORIL vs BC.} Using the same 189 demonstrations, varying the number of unlabeled episodes for an Insertion task with Jaco arm. Our method leverages unlabeled episodes and eventually achieves expert level.}
  \vspace{-0.5cm}
  \label{fig:unlabeled_data_improvement}
\end{wrapfigure}
Robotic manipulation poses several obstacles to contemporary deep learning approaches.
First, robotic interaction data may be costly and time-consuming to obtain, making online learning from scratch impractical.
Second, task rewards may be unobtainable or highly sparse, making conventional reinforcement learning a challenge.
Therefore, our objective is to develop an agent that can be trained offline from previously stored robot experience, as advocated by data-driven robotics approaches~\cite{cabi2020sketchy,fu2020d4rl,gulcehre2020rl}, without requiring reward annotations.

Behavior cloning (BC) agents can learn without task rewards by supervised learning on demonstration set.
However, as a result, BC does not effectively use non-expert data, since it may be sub-optimal or from a different task.
Offline RL could use all of the data, but only if rewards were provided.

Our proposed Offline Reinforced Imitation Learning (ORIL) agents combine the advantages of both, by the following:
(1)~learn a reward model by contrasting expert and unlabeled set observations, (2)~annotate all stored data using the learned model, and (3)~perform offline RL to learn a policy.
Hence, unlabeled data is used to train both the reward model and the policy and value function.
We observe that the performance improves as more unlabeled data is provided.

We find that our proposed approach benefits from incorporating components from several recently developed methods for adversarial imitation~\citep{zolna2020task}, positive-unlabeled reward modeling~\citep{xu2020positive} and a state-of-the-art approach to offline reinforcement learning~\citep{wang2020critic}.

As we show experimentally, ORIL has the following desirable characteristics.
\begin{itemize}
    \item Scalability -- performance improves as unlabeled data is added.
    \item Robustness -- works well even when the average quality of unlabeled data is low.
    \item Sample-efficiency -- outperforms BC with a much smaller demonstration set.
\end{itemize}

In summary, ORIL is an effective method for sample-efficient offline policy learning from demonstrations, from pixels and \emph{without} task rewards, that effectively leverages unlabeled experience.


\section{Method}
\label{sec:method}

At a high level, our method consists of the following steps: (1) learn a reward function by contrasting expert and unlabeled observations, (2) annotate both demonstrations and unlabeled trajectories with the learned reward, and then (3) perform offline RL.

Let $\mathcal{D}_{E}$ be the set of expert demonstrations and $\mathcal{D}_{U}$ be the set of unlabeled trajectories.
We assume that $\mathcal{D}_{U}$ can contain a mix of sub-optimal episodes for the targeted task, or behavior of another task, and also potentially some successful episodes, although proportionally fewer than in $\mathcal{D}_{E}$.
Let $\tau := (s_1, a_1, ..., s_T, a_T) \in \mathcal{D}_E \cup \mathcal{D}_U$ be an episode of length $T$ with observations $s_t$ and actions $a_t$.\footnote{We assume Markov Decision Process (MDP) and follow the usual RL notation of \citet{sutton2018reinforcement}.}
We assume that observations could be pixels or low-dimensional states such as joint angles.

\subsection{Learning a reward function}
\label{sec:reward_function}

We learn a reward model by contrasting expert and unlabeled trajectory states, similarly to how the discriminator is trained in Generative Adversarial Imitation Learning~(GAIL)~\citep{ho2016generative}.
However, we learn the reward function from a fixed dataset, unlike GAIL which requires access to the environment to generate new trajectories. Consequently, our method is especially suitable for robotic settings, where executing policies during learning can be too costly or slow.

The simplest approach, which we refer to as \emph{flat rewards}, is to label all states of expert demonstrations with a reward of 1, and all states of all unlabeled robot experience with a reward of 0.
Although this strategy is naive, it has been shown to be successful in online settings \cite{reddy2020}.

We can go beyond this hard-labeling approach by training a 
reward function $R_\psi(s_t)$ with parameters $\psi$ using the following cross-entropy loss:
\begin{align}
    \mathcal{L}_\psi(\mathcal{D}_E, \mathcal{D}_U) = \mathbb{E}_{s_t \sim \mathcal{D}_E} [-\log R_\psi(s_t)] + \mathbb{E}_{s'_t \sim \mathcal{D}_U} [-\log (1 - R_\psi(s'_t)],
    \label{eq:reward_obj}
\end{align}
where $s_t$ and $s'_t$ are sampled uniformly from all states in episodes in $\mathcal{D}_E$ and $\mathcal{D}_U$, respectively.

Note that the loss $\mathcal{L}_\psi$ is minimized when $R_\psi(s_t)$ assigns 1 to all expert observations and 0 to all unlabeled states, as in flat rewards.
This limiting behavior is unsatisfactory because unlabeled trajectories can contain both successful or unsuccessful states, without us having access to this information.
This existence of {\em false negatives} can hinder online adversarial imitation learning~\cite{zolna2020combating}.
To overcome this problem, we adapt two approaches for addressing false negatives, which have previously proved to be helpful in online adversarial learning: positive-unlabeled (PU) learning~\cite{elkan2008learning,xu2020positive} and Task-Relevant Adversarial Imitation Learning (TRAIL)~\cite{zolna2020task}.

\emph{Positive-unlabeled (PU) Learning} addresses the problem of false negatives directly~\cite{elkan2008learning}.
The main idea is to obtain an estimate of model loss on negative examples, which are not directly available, by re-weighting losses that can be computed using only positive and unlabeled data.
In our context, negative examples would correspond to observations not indicating a success condition for the task, while positive examples would indicate a success condition (e.g. stacking blocks in the correct order).
The PU Learning objective is described in the supplementary material; see~\cite{elkan2008learning,du2015convex,xu2020positive} for a derivation.

\emph{TRAIL} is an adversarial imitation learning method similar to GAIL, but in TRAIL the discriminator is constrained so as to \emph{not} form spurious associations between visual features and expert labels.
For example, the discriminator should not be able to distinguish whether early observations at the very start of an episode come from the demonstration or unlabeled set, since no meaningful behavior has yet been performed.
TRAIL has been shown to improve performance on robotic manipulation tasks and also address the problem of false negatives~\cite{zolna2020task}.

We adapt TRAIL to the offline setting. We use early observations\footnote{We define early observations as the first 10 observations in each episode.} to form TRAIL constraint sets (as proposed in the original paper).
Further details are included in the supplementary material.
We refer to the original paper for details~\cite{zolna2020task}.

\subsection{Reward model regularization}\label{sec:regularization}

\paragraph{Data augmentation}
We would like our reward model to generalize even when only a handful of episodes constitute $\mathcal{D}_E$.
However, given a set $\mathcal{D}_E$ that is very limited in size, a reward model can achieve perfect score by simply memorizing all these expert states and blindly assigning reward 0 to all the other states.
We applied extensive data augmentation to all reward model inputs during training (as in \cite{zolna2020task}), to alleviate this problem.

\paragraph{Training set split}
In the offline setting, we only have access to logged data and cannot interact with the environment.
Learning the reward model on the entire dataset, 
and subsequently using the model to annotate the same dataset, is prone to overfitting.
To overcome this, we split the unlabeled dataset in half and use only one half to train the reward model.
We learn the policy (including critic) on the full dataset.
We experimented with a few alternatives, but found this simple strategy to be sufficient for preventing overconfident reward estimates.

\subsection{Training an offline RL agent}

To train our offline RL agent, we use Critic-Regularized Regression (CRR)~\citep{wang2020critic}, a state-of-the-art offline reinforcement learning method.
The method assumes access to rewards and we alleviate that by using rewards issued by a reward model (trained as explained in Subsections~\ref{sec:reward_function} and~\ref{sec:regularization}).

CRR is advantageous because its policy update is simply a loss-weighted version of behavior cloning, where the per-example loss weights are determined by the critic network.
The critic, or state-action value function, is trained via Q-learning on the reward-annotated data.
In the most basic form of CRR, the policy loss is
\begin{align}
\mathcal{L}_\phi &= \mathbb{E}_{s_t, a_t \sim \mathcal{D}_E \cup \mathcal{D}_U}\big[- \log P(a_t | \pi_\phi(s_t)) \mathbb{I}\left[Q_\theta(s_t, a_t) > Q_\theta(s_t, \pi_\phi(s_t))\right]\big]\label{eq:crr},
\end{align}
where $\pi_\phi$ and $Q_\theta$ are the trained policy and critic parameterized by $\phi$ and $\theta$, respectively.

Note that the policy update in Equation~\ref{eq:crr} is identical to that of behavior cloning, but scaled by an indicator.
The indicator function $\mathbb{I}(\cdot)$ is equal to 1 if the stored action $a_t$ is judged to be better than the sampled policy action $a'_t \sim \pi_\phi(s_t)$ according to the current state action value function $Q_\theta$, and is equal to 0 otherwise.
As in~\citep{wang2020critic,barth2018distributed}, we train the critic together with the policy using distributional Q-learning, using the learned reward function explained in Subsections~\ref{sec:reward_function} and \ref{sec:regularization}.


\section{Experimental setup}

\subsection{Environments}\label{sec:environments}

\begin{figure}[t]
  \vspace{-0.2cm}
  \centering
  \includegraphics[height=0.34\linewidth]{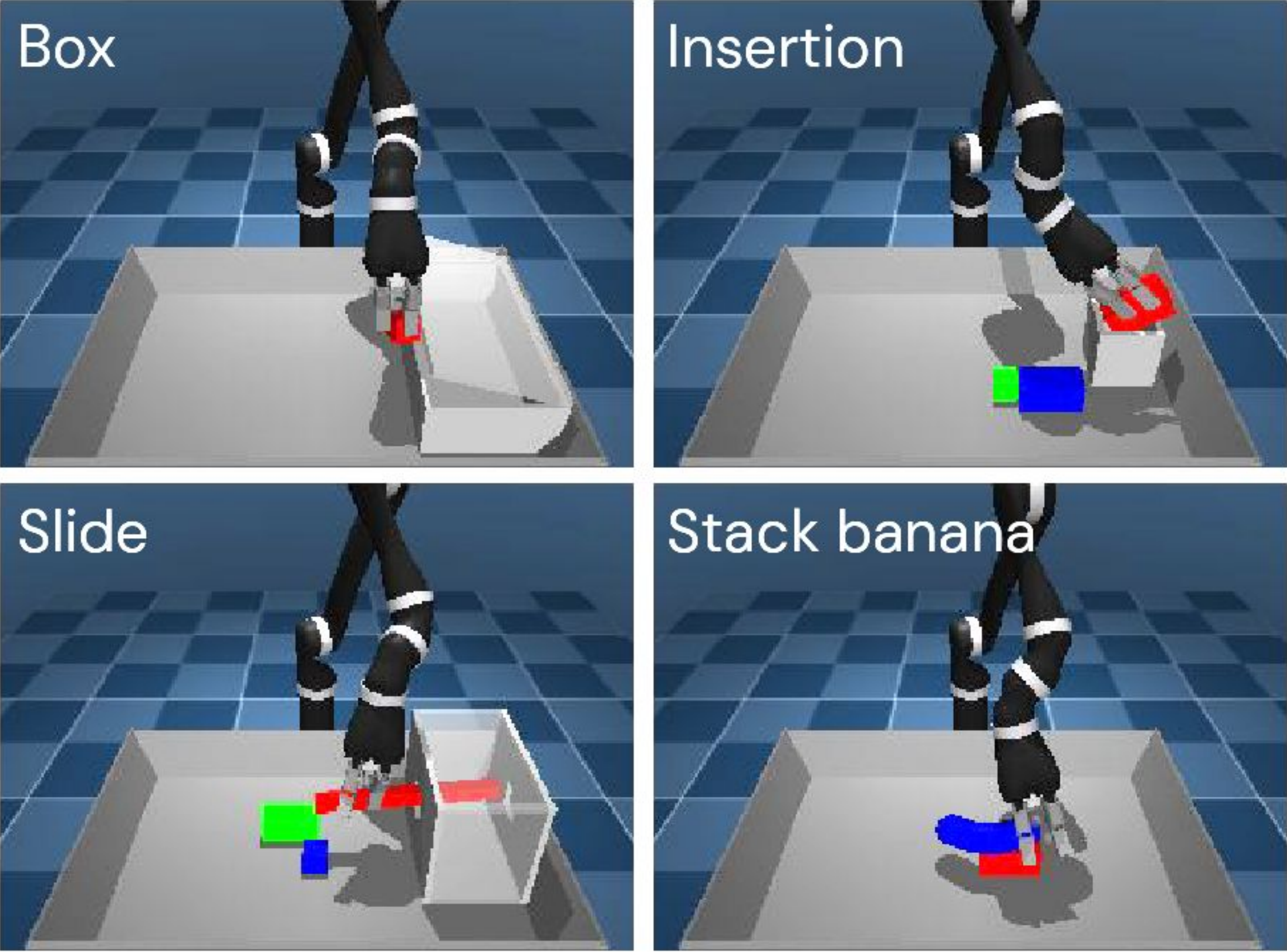}
  \hspace{0.35cm}
  \includegraphics[height=0.34\linewidth]{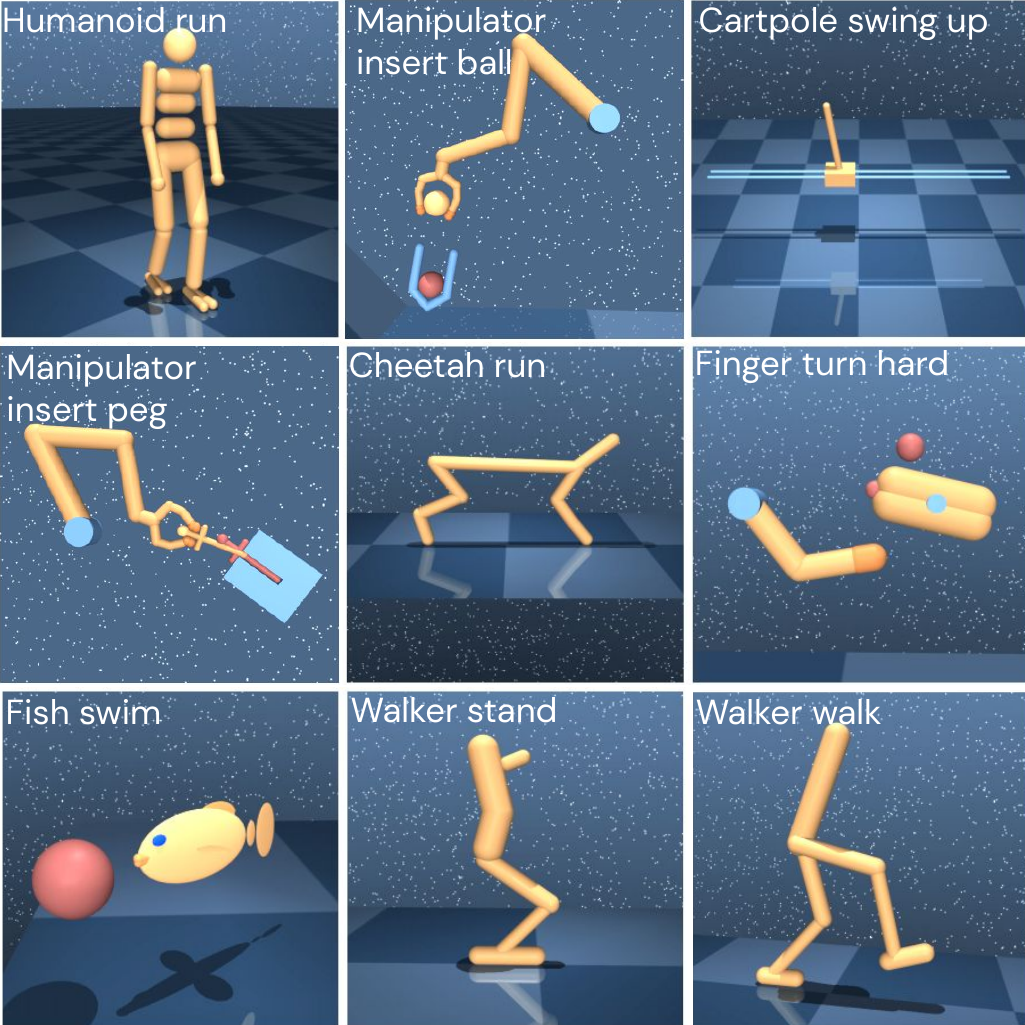}
  \caption{\textbf{Robotic Manipulation} (left) is a set of block manipulation tasks with a simulated Kinova Jaco arm in a 20x20cm basket.  \textbf{DeepMind Control Suite} (right) is a set of popular continuous control environments with tasks of varying difficulty, including locomotion and simple object manipulation.}
  \label{fig:domains}
  \vspace{-0.1cm}
\end{figure}

We conduct experiments on two simulated continuous control domains shown in Figure~\ref{fig:domains}: Robotic Manipulation and the DeepMind Control Suite.

\paragraph{Robotic Manipulation}

The first domain consists of four block manipulation tasks using a simulated Kinova Jaco arm in a 20cm by 20cm basket.
We use $64 \times 64$ pixel observations, and episodes are $400$ steps.
The environment reward is sparse ($1$ if task is solved and $0$ otherwise). 

\paragraph{DeepMind Control Suite}

The second domain is the DeepMind Control Suite, which contains a variety of continuous control tasks involving locomotion and simple manipulation.
Observations consist of joint angles and velocities, and action spaces vary depending on the task.
The episodes are $1000$ steps and the the environment reward is continuous, maximum $1$ per step.

\subsection{Datasets}\label{sec:datasets}
\begin{wraptable}{r}{0.41\textwidth}
\vspace{-0.45cm}
    \centering
    \caption{\textbf{Datasets statistics.} The total number of episodes (total) and corresponding number of demonstrations ($\#\mathcal{D}_E$) per task.}
    \vspace{0.1cm}
    \small
    \begin{tabular}{lrr}
    \toprule
    \textbf{Task} & \textbf{Total} & $\bm{\#\mathcal{D}_E}$ \\
    \midrule
    Box                      & 8000 & 161 \\
    Insertion                & 8000 & 189 \\
    Slide                    & 8000 & 189 \\
    Stack banana             & 8000 & 298 \\
    \midrule
    Cartpole swingup         &   40 &   2 \\
    Cheetah run              &  300 &   3 \\
    Finger turn hard         &  500 &   9 \\
    Fish swim                &  200 &   1 \\
    Humanoid run             & 3000 &  53 \\
    Manipulator insert ball  & 1500 &  30 \\
    Manipulator insert peg   & 1500 &  23 \\
    Walker stand             &  200 &   4 \\
    Walker walk              &  200 &   6 \\
    \bottomrule
    \end{tabular}
    \label{tab:episodes}
\vspace{-1.0cm}
\end{wraptable}
In practice, the set $\mathcal{D}_E$ would be collected by an expert and $\mathcal{D}_U$ is meant to be built based on the logged interactions of the agent from the past.

In this paper, we build these datasets based on existing offline RL datasets.
Given a set of logged episodes, we extract a small subset of well performing episodes and treat them as demonstrations $\mathcal{D}_E$.
The rest constitute the unlabeled set $\mathcal{D}_U$.
We use ground truth rewards only to perform this split and discard the rewards afterwards.
Table~\ref{tab:episodes} lists all tasks used in this paper and shows the total numbers of episodes and the sizes of expert sets for each task.
We explain the process for each domain below.

\paragraph{Robotic Manipulation}
The datasets for this environment are prepared as in~\cite{wang2020critic}.
There are 8000 episodes for each of four tasks and approximately 25\% of them are successful (i.e. at least half of the episode steps issue reward 1).
Each of these successful episodes was randomly chosen to be in $\mathcal{D}_E$ set with $\frac{1}{16}$ chance.

\paragraph{DeepMind Control Suite}
We use the open source RL Unplugged datasets~\cite{gulcehre2020rl}.
We define an episode as positive if its episodic reward is among top 20\% episodes for the task.
Each of these positives was randomly chosen to constitute $\mathcal{D}_E$ set with $\frac{1}{10}$ chance.
We note that resulting $\mathcal{D}_E$ sets are very small -- usually less than 10 episodes (see Table~\ref{tab:episodes}).

\subsection{Baseline and ablated models}

We implement two variants of behavioral cloning:

\textbf{BC}$\bm{_{pos}}$ \hspace{0.1cm}
Behavioral cloning on positive (expert) data only.
This baseline is trained only on $\mathcal{D}_E$ and hence, it is not exposed to low performing episodes in $\mathcal{D}_U$.
On the other hand, it may not be able to generalize due to the very limited size of its training set ($\mathcal{D}_E$).
\\%
\textbf{BC}$\bm{_{all}}$ \hspace{0.1cm}
Behavioral cloning on all data.
BC$_{all}$ may generalize better than BC$_{pos}$, due to access to a much larger dataset, but its performance may be worse if the quality of data in $\mathcal{D}_U$ is low.

Our method, ORIL, is composed of a batch RL agent (CRR~\citep{wang2020critic}) with a learned reward model.
To understand the contribution of each of its components we implement the following variants.

\textbf{FR} \hspace{0.1cm}
This method, similarly to SQIL~\cite{reddy2020}, assumes flat rewards.
It does not train a reward model and hence leverages only the CRR  component of our method.
\\%
\textbf{ORIL}$\bm{_{R}}$ \hspace{0.1cm}
This method uses a reward model with generic regularization (data augmentation and splitting) as described in Subsection~\ref{sec:regularization}.
It uses neither PU learning nor TRAIL.
\\%
\textbf{ORIL}$\bm{_{P}}$ and \textbf{ORIL}$\bm{_{T}}$ \hspace{0.1cm}
Improved ORIL$_{R}$ which additionally uses PU learning or TRAIL, respectively.

In following sections, when we use \textbf{ORIL} without subscript we mean ORIL$_{P}$ by default, since this variant performed best overall (see Subsection~\ref{sec:results}).

We also experimented with \textbf{off-policy GAIL}~\cite{kostrikov2018addressing} applied directly to our offline setting (with D4PG actor-critic~\cite{barth2018distributed}). However, the final returns were close to zero with a learnt reward function (suggesting that obtaining policy gradients via offline algorithm is crucial) and therefore we do not report detailed results.
Finally, we trained \textbf{CRR} with ground-truth rewards to obtain performance upper-bounds.

\subsection{Hyperparameters}

We used the same hyperparameters (including network architectures) as in CRR~\cite{wang2020critic}.
For the reward model, we used exactly the same network architecture as for the critic, but we replaced the final layer with a simple MLP predicting one value followed by sigmoid to scale its predictions.

We train all our models for $1e6$ learner steps and always with 3 seeds.
We compute the mean return between $5e5$ and $1e6$ learner steps for each seed.
This yields three numbers (one per seed) for every agent and we report mean and standard deviation across these numbers.


\section{Experimental results}
\label{sec:results}

\subsection{Robotic Manipulation}

\begin{figure}[ht]
  \vspace{-0.2cm}
  \centering
  \includegraphics[height=0.22\linewidth]{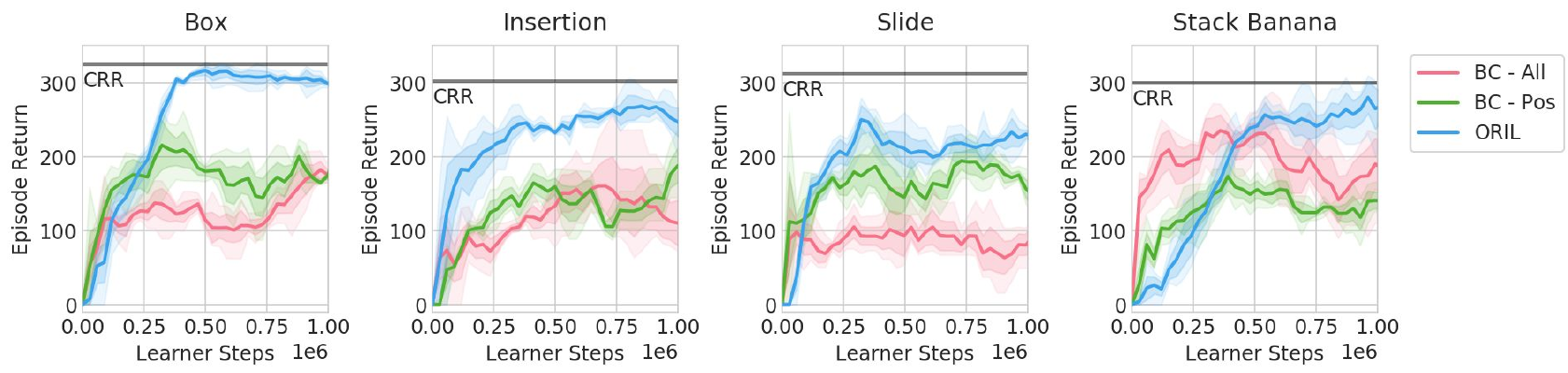}
  \caption{\textbf{Robotic Manipulation results.} We compare BC$_{all}$ (BC trained on all data), BC$_{pos}$ (BC trained only on demonstration data) and ORIL. ORIL improves over baselines by leveraging the unlabeled experience.}
  \label{fig:mpg}
  \vspace{-0.1cm}
\end{figure}

The learning curves for ORIL and both BC variants are shown in Figure~\ref{fig:mpg}.

ORIL closely approaches CRR upper-bound (trained with ground-truth rewards) and outperforms the baselines on all four tasks, showing that ORIL is well suited to make effective use of the unlabeled, mixed quality, trajectories.
There is no clear winner between BC$_{pos}$ and BC$_{all}$ which suggests that the quality of the unlabeled data for the considered tasks varies.
The results including ablated variants of ORIL are presented in Table~\ref{tab:ablations}.

\begin{table}[h]
    \vspace{-0.25cm}
    \centering
    \caption{\textbf{ORIL ablations on Robotic Manipulation domain.} We ablate and compare different variations of BC (BC$_{all}$ and BC$_{pos}$) and ORIL. For ORIL we compare, ORIL (the original method), ORIL$_{T}$ (using TRAIL instead of PU-learning) and ORIL$_{R}$ (with general purpose only regularization -- no PU learning nor TRAIL). Finally, we have FR which uses flat rewards (and hence does not train a reward model). We also provide per task performance upper-bounds obtained with CRR using ground-truth rewards (in italic).}
    \vspace{0.1cm}
    \small
    \begin{tabular}{l|llllll|l}
    \toprule
    \textbf{Task} & \multicolumn{1}{c}{\textbf{BC}$\bm{_{all}}$}  & \multicolumn{1}{c}{\textbf{BC}$\bm{_{pos}}$}  & \multicolumn{1}{c}{\textbf{ORIL}}    & \multicolumn{1}{c}{\textbf{ORIL}$\bm{_{T}}$} & \multicolumn{1}{c}{\textbf{ORIL}$\bm{_{R}}$} & \multicolumn{1}{c|}{\textbf{FR}} & \multicolumn{1}{c}{\textbf{\textit{CRR}}} \\
    \midrule
    Box  & 158 $\pm$  5   & 180 $\pm$  7   & \textbf{305} $\pm$  3 & \textbf{295} $\pm$  8  & \textbf{303} $\pm$  5  & \textbf{302} $\pm$  6  & \textit{325 $\pm$  4} \\   
    Insertion     & 146 $\pm$  8   & 139 $\pm$  5   & \textbf{260} $\pm$  3 & \textbf{266} $\pm$  3  & 241 $\pm$  4  & 100 $\pm$  4 & \textit{302 $\pm$ 12} \\  
    Slide         & 103 $\pm$  2   & 181 $\pm$  5   & \textbf{214} $\pm$ 13 & \textbf{224} $\pm$ 19  & 173 $\pm$ 13  & \mbox{ } 58 $\pm$  10 & \textit{312 $\pm$  9} \\  
    Stack Banana  & 210 $\pm$ 12   & 129 $\pm$  7   & \textbf{257} $\pm$  7 & 178 $\pm$ 25  & 232 $\pm$  8  & 208 $\pm$  5 & \textit{300 $\pm$  3} \\ 
    \bottomrule
    \end{tabular}
    \label{tab:ablations}
    \vspace{-0.1cm}
\end{table}

ORIL outperforms the other methods in all four tasks, showing 20-100\% higher returns than BC-based methods.
Our method works well with both PU learning (ORIL) and TRAIL (ORIL$_{T}$).
When only general purpose regularization is used (ORIL$_{R}$) the performance drops on all tasks but one.

FR preforms significantly worse than ORIL$_{R}$.
FR lags even behind the BC baselines on two tasks.
It primarily indicates that learning a reward model is helpful.
It also suggests that the reward model should be regularized, as FR effectively emulates a fully overfitted reward model which minimizes the loss $\mathcal{L}_\psi(\mathcal{D}_E, \mathcal{D}_U)$ in Equation~\ref{eq:reward_obj}.

\subsection{DeepMind Control Suite}

\begin{figure}[ht]
  \vspace{-0.2cm}
  \centering
  \includegraphics[height=0.4\linewidth]{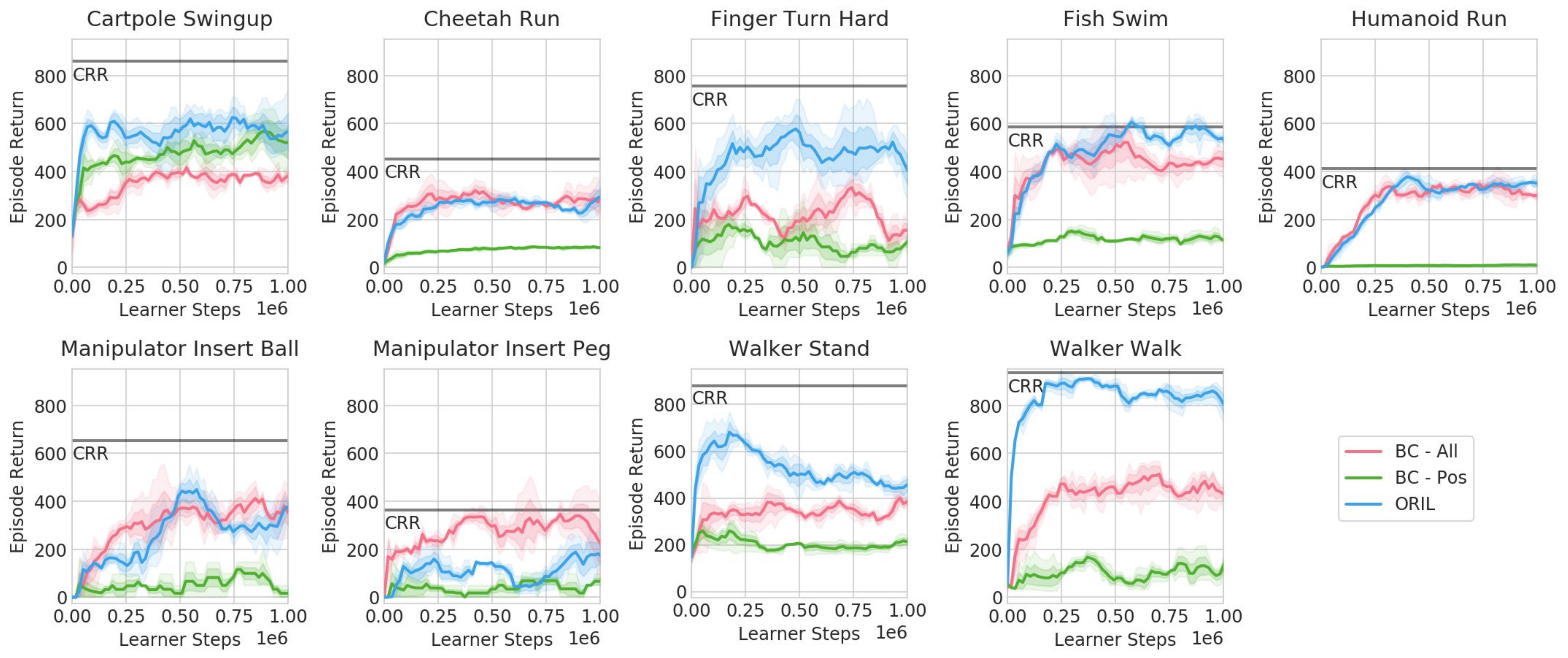}
  \caption{\textbf{DeepMind Control Suite results.} Our method (ORIL) is the best on 8 tasks (although BC$_{all}$ is equally good on 3 of them). BC$_{pos}$ is usually the worst due to the very limited set of demonstrations.}
  \label{fig:dmcs:results}
  \vspace{-0.1cm}
\end{figure}

We present results comparing our method and BC baselines in Figure~\ref{fig:dmcs:results}.

Our method outperforms the baselines on 5 out of 9 tasks.
For the other three tasks, the performance is comparable with BC$_{all}$, and ORIL is behind for only one task (Manipulator Insert Peg).
ORIL is able to reach (or closely approach) CRR upper-bounds even though our method does not need ground-truth rewards (see, e.g. Finger Turn Hard, Walker Walk).

BC$_{pos}$ performs poorly compared to other methods due to very small sizes of $\mathcal{D}_E$ for this domain.
It usually scores below 200.
Supervised learning on all data (BC$_{all}$) leads to significantly better results, suggesting that the data in $\mathcal{D}_U$ is of relatively high quality.
We also found that the performance of BC (but not of ORIL) is reduced when low-quality data is added to $\mathcal{D}_U$ (see Section~\ref{sec:low_performance_data}).

\section{Data set ablations}
\label{sec:additional:results}

\subsection{Varying the amount of unlabeled data}

\begin{figure}[h]
  \centering
  \vspace{-0.2cm}
  \includegraphics[height=0.33\linewidth]{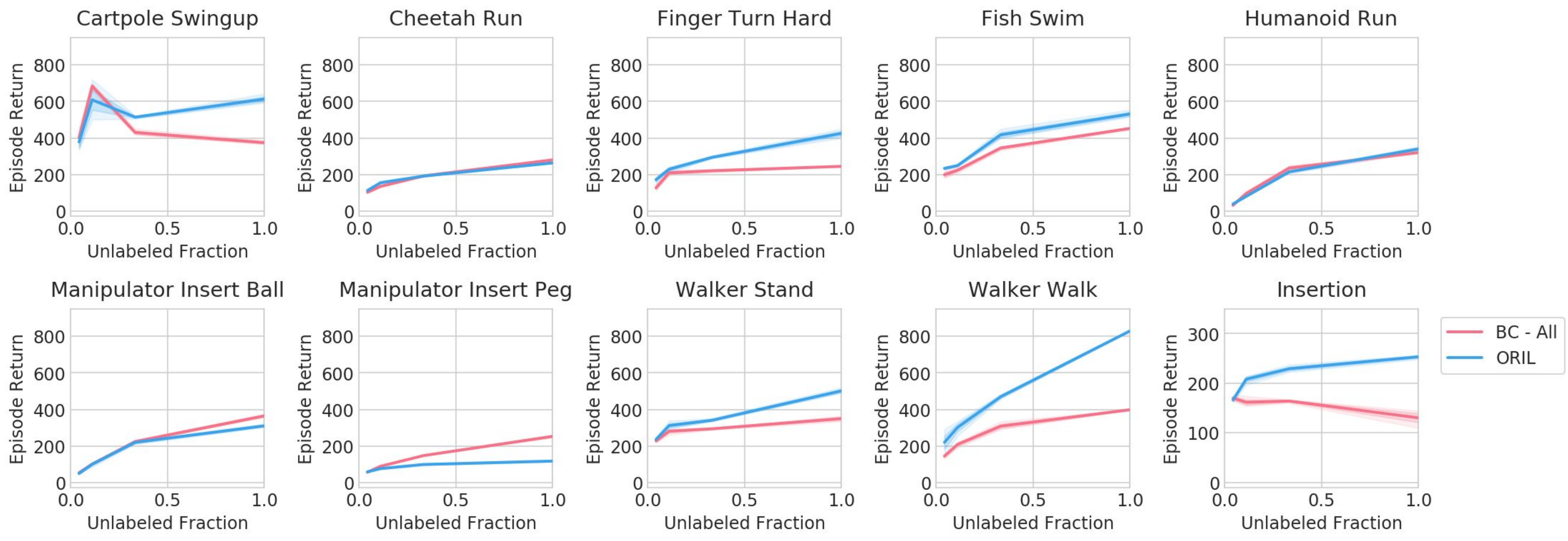}
  \caption{\textbf{Ablating the number of unlabeled trajectories.} We investigate the effect of unlabeled trajectories on agent performance. ORIL's performance clearly improves as the number of unlabeled data increases, whereas BC$_{all}$ either does not benefit from the extra data as much, or performs worse.}
  \label{fig:dmcs:var_unlabeled}
  \vspace{-0.1cm}
\end{figure}

In this experiment we vary the amount of unlabeled experience available for agent training.
Figure~\ref{fig:dmcs:var_unlabeled} shows that adding more unlabeled data improves the performance of ORIL, whereas BC does not benefit from the extra data as consistently, and sometimes performs worse.

\subsection{Adding low performance data to the unlabeled set}
\label{sec:low_performance_data}

The average data quality for DeepMind Control Suite tasks is relatively high, as evidenced by the fact that BC$_{all}$ performs better than BC$_{pos}$.
We conducted a set of experiments to check if BC$_{all}$ and our method are robust when the ratio of low performing episodes in $\mathcal{D}_U$ increases.
We doubled the sizes of the original datasets for four tasks by adding episodes generated only by low performing policies.
We kept $\mathcal{D}_E$ as before, and hence all the extra episodes landed in $\mathcal{D}_U$.

We consider four tasks: Finger Turn Hard and Walker Walk (where our method is clearly better than the baselines on the original set), and Fish Swim and Humanoid Run (where the difference is smaller).
We report average evaluation scores for the original and amended datasets in Table~\ref{tab:unlabeled}.

\begin{table}[h]
\vspace{-0.25cm}
    \caption{\textbf{Robustness to the low-quality data.} ORIL is more robust than BC to adding low-quality data that has episodes with low average rewards (indicated by ``Amended'').\label{tab:unlabeled}}
    \vspace{0.1cm}
    \centering
    \small
    \begin{tabular}{ll|l|l}
    \toprule
    & & \multicolumn{2}{c}{\textbf{Scores}} \\
    \textbf{Environment} & \textbf{Dataset} & \multicolumn{1}{c}{\textbf{ORIL}} & \multicolumn{1}{c}{\textbf{BC}$\bm{_{all}}$} \\
    \midrule
    \multirow{2}{*}{Finger Turn Hard} & Original   & \textbf{406} $\pm$ 11.12 & 250 $\pm$ 7.31 \\
                                      & Amended    & \textbf{446} $\pm$ 17.79 (+10\%)& 200 $\pm$ 1.18 (-20\%) \\
    \midrule
    \multirow{2}{*}{Fish Swim}        & Original   & \textbf{518} $\pm$ 10.43 & 458 $\pm$ 5.15 \\
                                      & Amended    & \textbf{448} $\pm$  3.76 (-13\%) & 349 $\pm$ 4.07 (-23\%) \\
    \midrule
    \multirow{2}{*}{Humanoid Run}     & Original   & \textbf{349} $\pm$  2.51 & 296 $\pm$ 25.91 \\
                                      & Amended    & \textbf{254} $\pm$  5.48 (-27\%) & 217 $\pm$ 3.74 (-26\%) \\
    \midrule
    \multirow{2}{*}{Walker Walk}      & Original   & \textbf{829} $\pm$ 15.43 & 396 $\pm$ 5.70 \\
                                      & Amended    & \textbf{836} $\pm$  5.00 (+1\%) & 248 $\pm$ 8.16 (-37\%) \\
    \bottomrule
    \end{tabular}
\vspace{-0.1cm}
\end{table}

As expected, BC$_{all}$ always performs worse when exposed to low-quality episodes.
The largest performance drop is for Walker Walk which is 37\%.

ORIL is shown to be significantly more robust.
For two tasks, there is no drop in performance at all, and there is even a small improvement for Finger Turn Hard.
For the other two tasks the additional low-quality data negatively influence ORIL performance.
However, ORIL still outperforms the baseline significantly.
It confirms that CRR (the offline RL agent) using learned rewards is able to filter out transitions which would reduce performance if trained on (see Equation~\ref{eq:crr}).

\subsection{Varying the number of demonstration episodes}

As mentioned in Subsection~\ref{sec:datasets}, only 10\% of positive episodes are used to constitute demonstration sets for DeepMind Control Suite tasks.
As presented in our main experiments, ORIL is always significantly better than BC$_{pos}$ in this challenging setting (see Section~\ref{sec:results}) which indicates the much better sample complexity of our method.

\begin{figure}[ht]
  \vspace{-0.2cm}
  \centering
  \includegraphics[height=0.4\linewidth]{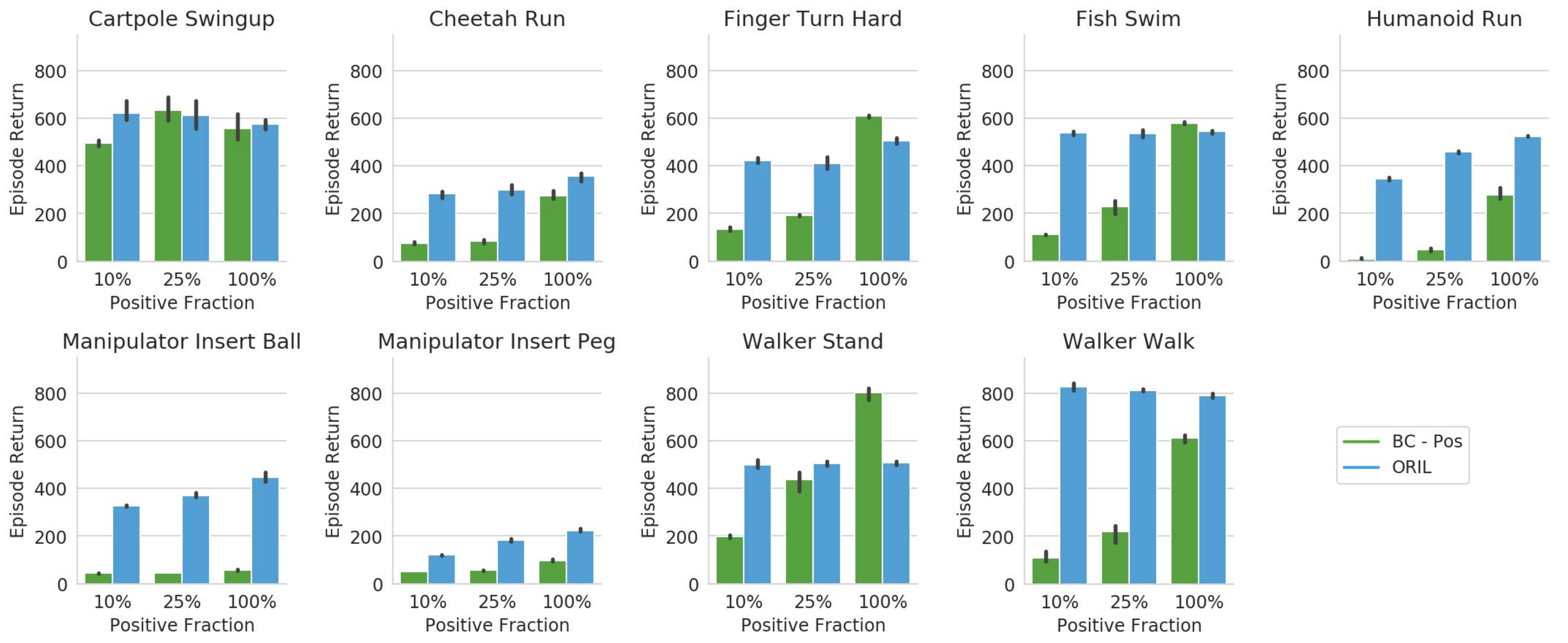} 
  \caption{\textbf{Control suite results.} ORIL is more robust to decreasing the amount of demonstrations than BC$_{pos}$.}
  \label{fig:dmcs:var_demos}
  \vspace{-0.1cm}
\end{figure}

Here, we analyze if BC$_{pos}$ can reach ORIL performance when more (25\%) or even all positives are used to build $\mathcal{D}_E$.
Figure~\ref{fig:dmcs:var_demos} shows the effect of increasing the proportion of demonstrations.

BC$_{pos}$ achieves better performance when more positives are used, but still performs worse than ORIL.
It indicates that ORIL is able to leverage signal present even in low-quality episodes.

ORIL maintains its performance for all three versions of $\mathcal{D}_E$ sets.
It suggests that the reward model can generalize well to unlabeled data even when trained with a very limited number of demonstrations.
We note that sizes of demonstrations sets are usually below 10 for the tasks considered.

\section{Related work}
\label{sec:related_work}

Imitation is an important topic in the study of intelligence and consequently there is a vast body of literature on this topic. \citet{osa2018an} present a comprehensive and authoritative review of imitation in robotics and machine learning. We refer readers to this review and provide a brief summary next.

The most obvious way to imitate is to mimic exactly what another agent is doing via supervised learning. This is often referred to as behaviour cloning (BC)~\citep{pomerleau1989alvinn}. As with all supervised learning methods, BC works best when large datasets of high quality demonstrations are available. This high sample complexity is caused by the fact that BC performs poorly in states that are far away from the training data, though some techniques have been proposed to mitigate this problem \cite{ross2011reduction}. Moreover, since the focus of BC is on mimicking, it will not learn to supersede demonstrators or learn to select what to imitate to achieve a goal more effectively.

Inverse RL~\citep{ng2000algorithms, abbeel2004apprenticeship} provides an alternative to BC. The focus is on learning a reward function from high-quality expert trajectories, and using this reward function to learn a policy. 
Recent examples of this approach \cite{finn2016guided, ho2016generative,li2017infogail,fu2017learning,merel2017learning,zhu2018reinforcement,baram2017end}, have been particularly successful when learning from states. However, learning effectively from raw input video streams remains a challenge, see for example~\cite{zolna2020task}. These approaches typically require the online execution of RL agents for many trials, making them too costly for real robots. In contrast, in this paper we learn the rewards and policies directly from logged data, obviating the need for online execution during learning.

Demonstrations are often used to improve RL agents by guiding exploration~\cite{nair2018overcoming, rajeswaran2017learning,vecerik2017leveraging,pohlen2018observe, vecerik2019practical,zolna2019reinforced,paine2019making}. This is often achieved by replaying these demonstrations, alongside recent online experiences, in model-free RL approaches. While we use demonstrations, we emphasize again that we learn purely from logged data, and without online deployment during learning.

Learning from positive and unlabeled data is a practical challenge in many applications, such as medical diagnosis and robotics~\cite{xu2020positive}.
PU learning approaches aim at solving this problem, and the idea of unbiased risk estimators was a breakthrough~\cite{elkan2008learning,du2015convex,kiryo2017positive}.
\citet{xu2020positive} successfully applied PU learning in adversarial imitation and supervised reward learning.

Offline RL, often referred to as batch RL, has been reviewed by
\citet{lange2012batch} and more recently by
\citet{levine2020offline}. Many algorithmic variants have beep proposed over the last two years, including among others BCQ~\citep{fujimoto2019off}, MARWIL~\citep{wang2018exponentially}, BAIL~\citep{chen2020bail}, ABM~\citep{siegel2020keep} AWR~\citep{peng2019advantage}, and CRR~\citep{wang2020critic}. There has also been progress in benchmarking and offline hyper-parameter selection \cite{gulcehre2020rl,paine2020hyperparameter}. 

Most offline RL works assume that rewards are given. 
The work of~\citet{cabi2020sketchy} instead use a preference elicitation interface allowing humans to sketch reward curves, in order to learn reward functions.
Similar to our case, these learned rewards are then provided to an offline RL agent (D4PG~\cite{barth2018distributed} in the case of ~\citep{cabi2020sketchy} versus CRR~\citep{wang2020critic} in this work).
Our method differs mainly in that we do not need human annotators in the reward learning process, apart from potentially producing demonstration data.
Also, our reward learning method uses all available demonstration and unlabeled set data, whereas~\citep{cabi2020sketchy} only learn rewards from sketched episodes.

\section{Conclusion}
\label{sec:conclusion}

We proposed offline reinforced imitation learning (ORIL) to enable learning from both demonstrations and a large unlabeled set of experiences without reward annotations.
Leveraging unlabeled data allows ORIL to consistently outperform comparable BC agents.
Given a modest number of demonstrations, adding more unlabeled experience improves the performance of ORIL across a diverse set of continuous control and simulated robotic manipulation tasks.
We showed that ORIL is competitive with state-of-the-art methods that use ground-truth rewards, while ORIL itself does \emph{not} rely on ground-truth rewards.

{
\fontsize{9.5}{11}\selectfont
\bibliography{main}
\bibliographystyle{plainnat}
}

\clearpage
\appendix
{\Large \bf Supplementary material}
\normalsize

\section{Training procedure details}

\subsection{Algorithm details}

We train a policy, critic and reward model simultaneously as shown in Algorithm~\ref{alg:oril}.
However, the reward model can be trained separately first, and then used to annotate datasets $\mathcal{D}_E$ and $\mathcal{D}_U$ with reward estimates.
Both these approaches result in similar agent performance.

Algorithm~\ref{alg:oril} implements ORIL with PU learning (ORIL$_P$).
The data augmentation is done as in~\cite{zolna2020task}.
The critic update is based on the discrepancy between the current estimate of the action-value and the TD-update (where the next step reward is estimated by reward model $R_\psi$).
This discrepancy is measured by a divergence measure or a metric $D$, such as KL-divergence or squared error.
In this paper, we use a distributional Q-function as in~\cite{wang2020critic} and use the divergence measure proposed in~\cite{barth-maron2018distributional}.

\begin{algorithm}[h]

\SetAlgoLined
\small 
\textbf{Input:} critic, policy and reward networks: $Q_{\theta}$, $\pi_{\phi}$ and $R_\psi$, and its target versions: $Q_{\theta'}$, $\pi_{\phi'}$, $R_{\psi'}$, divergence measure $D$, expert dataset $\mathcal{D}_E$ and unlabeled dataset $\mathcal{D}_U$, hyperparameter $\eta$, number of updates $n_{updates}$\;
\mbox{ }\\
\textbf{Begin:} Split $\mathcal{D}_U$ in half to get $\mathcal{D}_U^1$ and $\mathcal{D}_U^2$.\\
\For{$n_{updates}$}{
    \tcc{Reward learning}
    Sample expert and unlabeled batches, i.e. $\mathcal{B}_E \subset \mathcal{D}_E$ and $\mathcal{B}_U \subset \mathcal{D}_U^1$\;

    Augment images in $\mathcal{B}_E$ and $\mathcal{B}_U$ to obtain augmented batches $\overline{\mathcal{B}_E}$ and $\overline{\mathcal{B}_U}$\;

    \SetKwBlock{TR}{\textnormal{Train reward model using augmented batches $\overline{\mathcal{B}_E}$ and $\overline{\mathcal{B}_U}$:}}{}
    \TR{
        Update reward network parameters $\psi$ with gradient:\\
        \hspace{0.2cm} $-\nabla_{\psi} \eta \mathbb{E}_{s_t \sim \overline{\mathcal{B}_E}}[-\log(R_{\psi}(s_t))] +  \mathbb{E}_{s_t \sim \overline{\mathcal{B}_U}}[-\log(1-R_{\psi}(s_t))] - \eta\mathbb{E}_{s_t \sim \overline{\mathcal{B}_E}}[-\log(1-R_{\psi}(s_t))]$
    }

    \mbox{ }\\
    \mbox{ }\\
    \tcc{Policy and critic learning}
    Sample additional unlabeled batch from $\mathcal{D}_U^2$, i.e. $\mathcal{B}_U^2 \subset \mathcal{D}_U^2$\;
  
    Concatenate (not augmented) expert and both unlabeled batches: $\mathcal{B} = \mathcal{B}_E \cup \mathcal{B}_U^1 \cup \mathcal{B}_U^2$\;
  
    \SetKwBlock{CRR}{\textnormal{Apply CRR updates using rewards predicted by $R_{\psi'}$:}}{}
    \CRR{
        Update critic network parameters $\theta$ with gradient:\\
        \hspace{0.2cm} $-\nabla_{\theta} \mathbb{E}_{(s_t, a_t, s_{t+1}) \sim {\mathcal{B}}}
        D\big[
            Q_{\theta}(s_t, a_t),
            R_{\psi'}(s_{t+1}) + \gamma\mathbb{E}_{a \sim \pi_{\phi'}(s_{t+1})} Q_{\theta'}(s_{t+1}, a)
        \big]$\;
        
        Update policy network parameters $\phi$ with gradient:\\
        \hspace{0.2cm} $-\nabla_{\phi} \mathbb{E}_{(s_t, a_t) \sim {\mathcal{B}}} \log \pi_{\phi} (a_t | s_t) \mathbb{I}\left[Q_\theta(s_t, a_t) > Q_\theta(s_t, \pi_\phi(s_t))\right]$\;
        }
    
    \mbox{ }\\
    \mbox{ }\\
    \tcc{Target networks update}
    Update the target networks every $N$ steps by copying parameters: $\theta' \leftarrow  \theta, \ \ \phi' \leftarrow  \phi, \ \psi' \leftarrow  \psi$\;
}
\caption{\small \textbf{Offline Reinforced Imitation Learning} \label{alg:oril}}
\end{algorithm}

\subsection{Hyperparameters}

\begin{figure*}[t]
\centering
\includegraphics[width=1\textwidth]{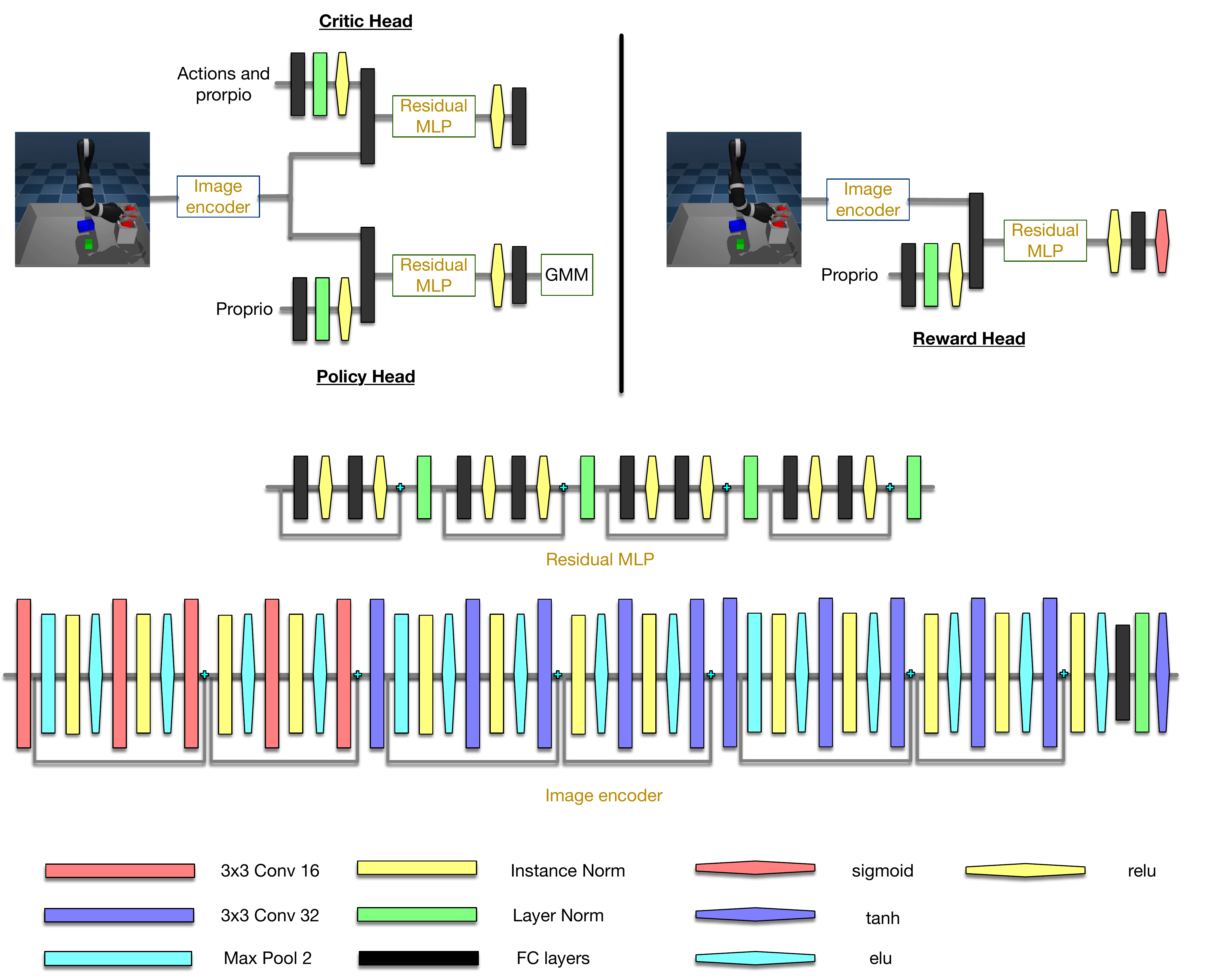}
\vspace{0.3cm}
\caption{\label{fig:architectures} \textbf{The architectures used by the critic, policy and reward models.}
There are two instances of the image encoder, one is shared between the critic and policy networks while the other is part of the reward network.
All models implement the same residual MLP but it is never shared.
GMM stands for Gaussian Mixture Model which outputs the final policy prediction (see details in the main text).
This figure is based on the architecture figure from CRR paper~\cite{wang2020critic}.
}
\end{figure*}

We parametrize the critic, policy and reward models with neural networks.
The policy and critic architectures are identical to the ones used in Critic Regularized Regression work \cite{wang2020critic}, and the reward model architecture is inspired by the critic network.
All networks are described in details below and presented in Figure~\ref{fig:architectures}.

We use the same hyperparameters as in \cite{wang2020critic}.
We train all our models for $1e6$ learner steps and always with 3 seeds.
We compute the mean return between $5e5$ and $1e6$ learner steps for each seed.
This gives three values of mean returns (one per seed) for every agent and we report mean, and standard deviation across these results.

For pixel based tasks (i.e. Robotic Manipulation tasks), pixels are encoded with a residual CNN (see Figure~\ref{fig:architectures}, bottom).
Two separate image encoders are trained.
One of them is shared between critic and policy networks, and another is used (and trained) solely by the reward model.

All networks process proprio states (for critic, they are concatenated with actions) with one layer MLP to obtain preliminary representations which are then concatenated with image representations (for pixel based tasks) and further processed.
The final layers depend on the network purpose (see details below).

\paragraph{Policy head}

Proprioceptive input is processed with a fully connected layer, layer normalization and tanh activation function, then concatenated with pixel encoding (when present) and passed into a residual MLP (see Figure~\ref{fig:architectures}, top left). The output of the MLP defines the policy which is a mixture of five multivariate Gaussians. Specifically, the output of the MLP is contains: five mean vectors; five vectors that (after passing through a softplus function) define diagonal of the covariance matrix of each Gaussian; and five scalars that define mixture log-probabilities.

\paragraph{Critic head}

Proprioceptive input concatenated with actions is processed through a fully connected layer, layer normalization and tanh, then concatenated with pixel encoding (when present) and passed into a residual MLP (see Figure~\ref{fig:architectures}, top left). The output of the MLP defines a discrete distribution of the distributional critic.

\paragraph{Reward head}

Proprioceptive input is processed through a fully connected layer, layer normalization and tanh, then concatenated with pixel encoding (when present) and passed into a residual MLP (see Figure~\ref{fig:architectures}, top right).
The output is proccesed by a fully connected layer to obtain a scalar which is then scaled to $(0,1)$ by applying sigmoid.

\section{Positive-unlabeled learning}
If we treat reward learning as a binary decision problem to distinguish success ($\mathcal{D}_E$) and failure ($\mathcal{D}_F$) trajectories, we can write the loss for the reward model $R$ as
\begin{align}
    \eta \mathbb{E}_{s_t \sim \mathcal{D}_E}[-\log(R(s_t))] + (1-\eta)\mathbb{E}_{s_t \sim \mathcal{D}_F}[-\log(1-R(s_t))],
\end{align}
where $\eta$ is the proportion of the trajectory space corresponding to success. This corresponds to a more general form of Equation~\ref{eq:reward_obj} where before $\eta = 0.5$ was assumed.

The key insight of PU-learning \cite{elkan2008learning,du2015convex} is to observe that the second term in the loss (computed for $\mathcal{D}_F$) can be written in terms of a dataset of unlabeled trajectories $\mathcal{D}_U$ (that contains an unknown mixture of successes and failures) and the dataset of positive trajectories $\mathcal{D}_E$:
\begin{align}
    (1-\eta)\mathbb{E}_{s_t \sim \mathcal{D}_F}[-\log(1-R(s_t))] = \mathbb{E}_{s_t \sim \mathcal{D}_U}[-\log(1-R(s_t))] - \eta\mathbb{E}_{s_t \sim \mathcal{D}_E}[-\log(1-R(s_t))].
\end{align}
This allows the loss for the reward model to be rewritten in a way that avoids any dependence on explicitly labeled failures
\begin{align}
    \eta \mathbb{E}_{s_t \sim \mathcal{D}_E}[-\log(R(s_t))] +  \mathbb{E}_{s_t \sim \mathcal{D}_U}[-\log(1-R(s_t))] - \eta\mathbb{E}_{s_t \sim \mathcal{D}_E}[-\log(1-R(s_t))]
\end{align}
In this paper we treat $\eta$ as a hyperparameter and set it to $\eta=0.5$ throughout.

\section{Task-relevant adversarial imitation learning}
TRAIL proposes to constrain the GAIL discriminator such that it is \emph{not} able to distinguish between certain, preselected expert and agent observations which do not contain task behavior.
For example, the discriminator should not be able to distinguish whether early observations at the very start of an episode come from the demonstration or unlabeled set, since no meaningful behavior has yet been performed.

We adapt TRAIL to the offline setting and use early observations to form our constraint sets.
Specifically, we construct subsets of early states, i.e. $\mathcal{D}_{U}' = \{ s_t \in \mathcal{D}_U \mid t < 10\}$, and analogously $\mathcal{D}_{E}'$ for $\mathcal{D}_E$.
We compute reward loss as in Equation~\ref{eq:reward_obj} for the original datasets and also, separately, for the states from the constraint sets.
The loss optimized by the reward model is the following:
\begin{equation}
    L_\psi(\mathcal{D}_E, \mathcal{D}_U) - \1_{R_\psi(\mathcal{D}_{U}') > R_\psi(\mathcal{D}_{E}')} L_\psi(\mathcal{D}_{E}', \mathcal{D}_{U}'),
\end{equation}
where $\1_{R_\psi(\mathcal{D}_{E}') > R_\psi(\mathcal{D}_{U}')}$ is one if average prediction $R_\psi$ for expert early observations ($\mathcal{D}_{E}'$) is higher than for unlabeled early observations ($\mathcal{D}_{U}'$).
Intuitively, we use early observations to first control if the reward model is overfitting, and if so, we use them again to compute reversed loss to regularize discriminator.
We refer to the original paper~\cite{zolna2020task} for motivation and detailed description.

The original paper~\cite{zolna2020task} proposes another indicator for applying reversed loss, but we find the average prediction for early observations used by us to work similarly well.

\end{document}